\documentclass[letterpaper, 10 pt, conference]{ieeeconf}  

\IEEEoverridecommandlockouts                              

\usepackage{cite}
\usepackage{relsize}
\usepackage{dsfont}
\usepackage{bbm}
\usepackage{color}

\usepackage{subfigure}
\usepackage[pdftex]{graphicx}
\graphicspath{ {pics/} }  
\DeclareGraphicsExtensions{.pdf,.jpeg,.png}

\usepackage{epstopdf}
\usepackage{epsfig}
\usepackage{color}

\usepackage{amsmath}
\usepackage{amsfonts}
\usepackage{amssymb}
\usepackage{bm}

\usepackage{txfonts}
\DeclareMathOperator*{\R}{\varmathbb{R}}

\usepackage{algorithm}
\usepackage{algpseudocode}
\usepackage{etoolbox}

\input{macro}

\title{\LARGE \bf 
Co-active Learning to Adapt Humanoid Movement for Manipulation
}

\author{\authorblockN{Ren Mao$^{1}$, John S. Baras$^{1}$, Yezhou Yang$^{2}$, and Cornelia Ferm$\ddot{\text{u}}$ller$^{2}$
\thanks{$^{1}$R. Mao and J. Baras are with the Department of Electrical and Computer Engineering and the ISR, $^{2}$Y. Yang and C. Ferm$\ddot{\text{u}}$ller are with the Department of Computer Science and the UMIACS, University of Maryland, College Park, Maryland 20742, USA. \{neroam, baras\} at umd.edu, yzyang at cs.umd.edu and fer at umiacs.umd.edu.} }}%

\begin{document}

\maketitle
\thispagestyle{empty}
\pagestyle{empty}

\begin{abstract}
In this paper we address the problem of robot movement adaptation under various environmental constraints interactively. Motion primitives are generally adopted to generate target motion from demonstrations. However, their generalization capability is weak while facing novel environments. Additionally, traditional motion generation methods do not consider the versatile constraints from various users, tasks, and environments. In this work, we propose a co-active learning framework for learning to adapt robot end-effector's movement for manipulation tasks. It is designed to adapt the original imitation trajectories, which are learned from demonstrations, to novel situations with various constraints. The framework also considers user's feedback towards the adapted trajectories, and it learns to adapt movement through human-in-the-loop interactions. The implemented system generalizes trained motion primitives to various situations with different constraints considering user preferences. Experiments on a humanoid platform validate the effectiveness of our approach.
\end{abstract}

\section{Introduction}

Trajectories learning from human demonstrations has been studied in the field of Robotics for decades due to its wide range of applications, in both industrial and domestic scenarios. Among the various approaches, Motion Primitives (MPs) aims to parameterize observed human motion and then reproduce it, given different initial and target states. However, it is widely known that general MPs methods, such as Dynamic Movement Primitives (DMPs) ~\cite{Ijspeert13}, have limited capability to generalize towards novel environments involving other constraints. Moreover, standard MPs learning method ignores user preferences of the tasks and the environments. For real world humanoid applications, a practical robot movement learning framework needs to take user preferences and environment constraints into consideration.

\begin{figure}[thpb]
\centering
  \includegraphics[width=0.9\columnwidth]{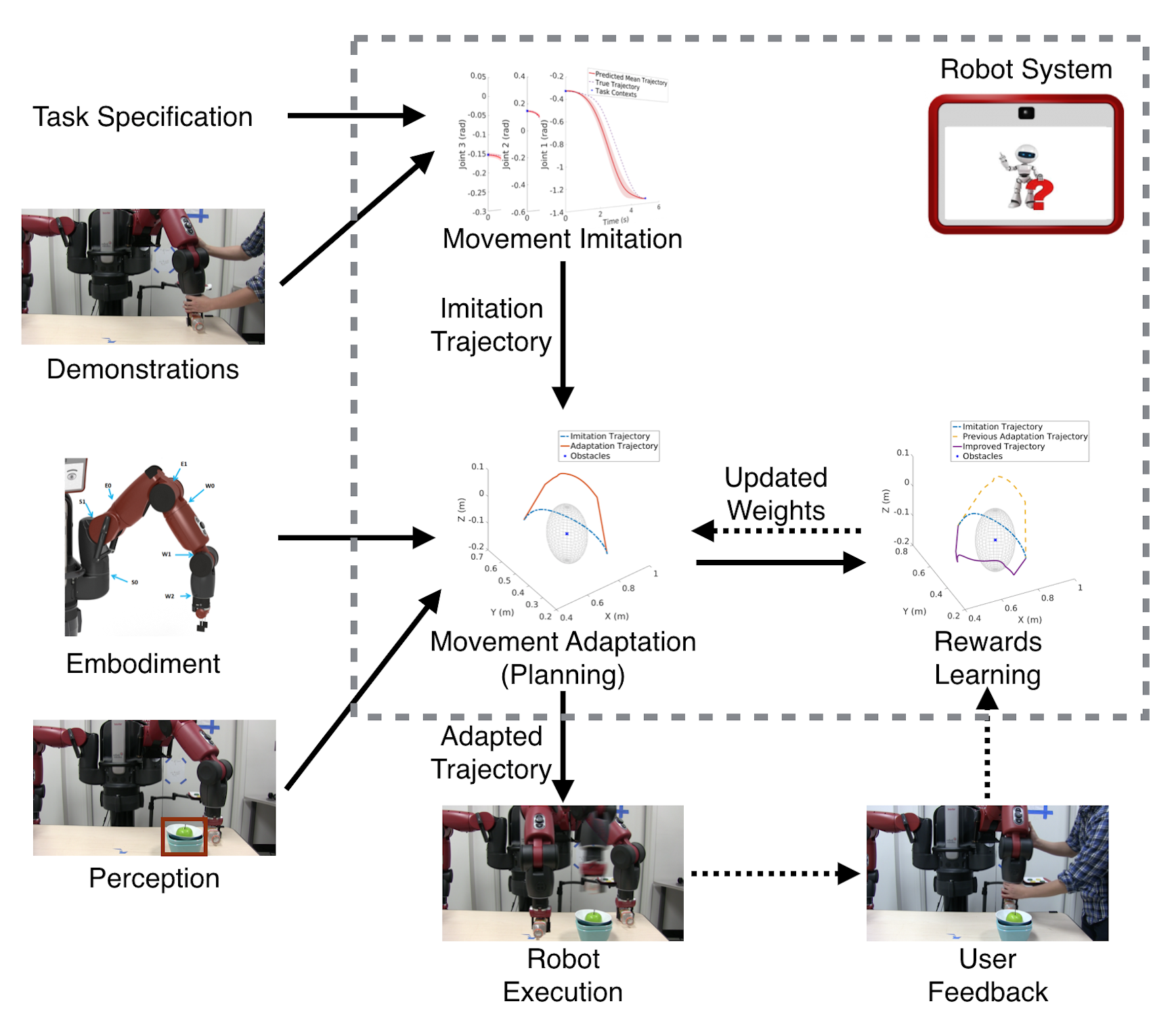} 
  \caption{System for learning movement adaptation for manipulation tasks. Dashed lines indicate feedback.}
  \label{fig:Block_diagram_adapt}
  \vspace{-12pt}
\end{figure}

\begin{figure*}[thpb]
  \subfigure[]{
    \centering
    \includegraphics[height=1.2in]{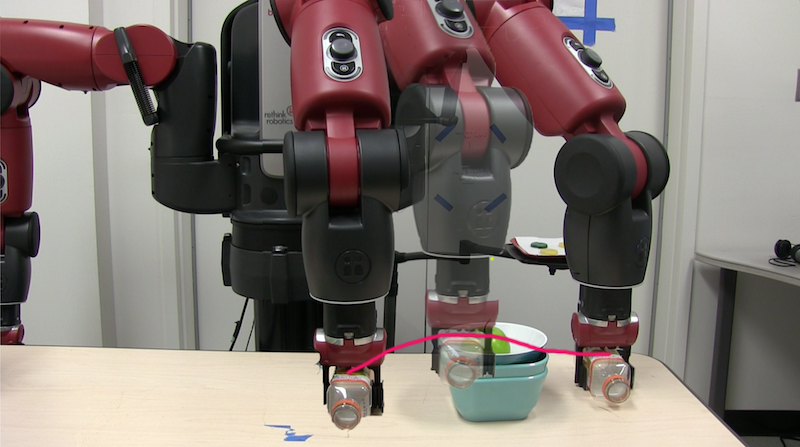} 
    \label{fig:baxterImitation}
  }
  \subfigure[]{
    \centering
    \includegraphics[height=1.2in]{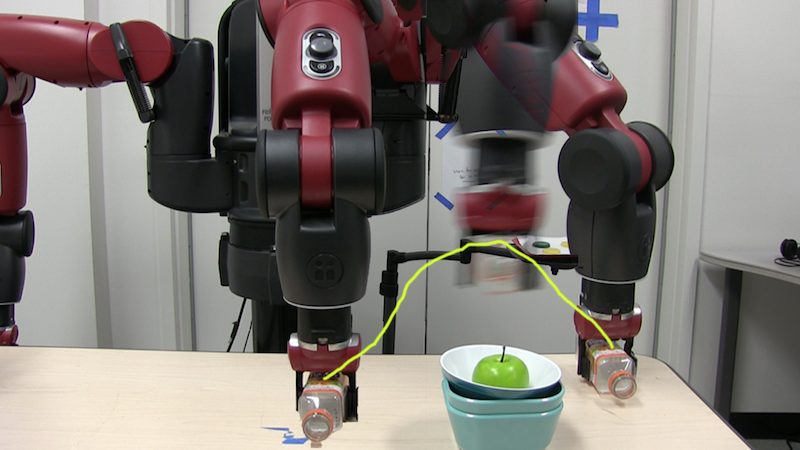} 
    \label{fig:baxterAdaptation}
  }
  \subfigure[]{
    \centering
    \includegraphics[height=1.2in]{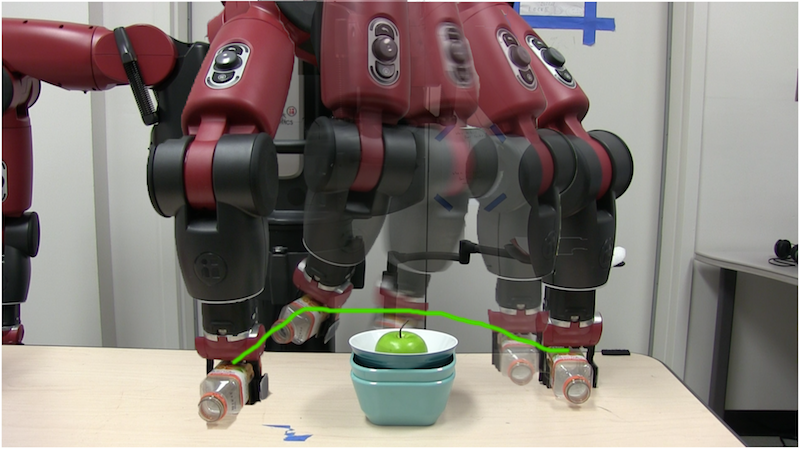} 
    \label{fig:baxterLearnAdaptation}
  }
  \caption{Baxter Transferring Leaking Bottle: (a) Movement imitation, failed to avoid the bowl; (b) Movement adaptation with initial weights, successfully avoided the bowl by path above it but spilled water in the bowl; (c) Movement adaptation with learned weights for new situation where obstacle locates differently, successfully avoided the bowl by path around it and avoided spilling water in the bowl.}
  \label{fig:BaxterExecution}
  \vspace{-12pt}
\end{figure*}

Let's start from a common example, that a human user teaches a humanoid how to transfer a bottle from different start and end states. Using off-the-shelf approach, the robot is able to learn the motion by acquiring MPs from the demonstrated trajectories and apply them to generate new trajectories given different initial and target states. However, solely following the generated trajectories may fail if the environment has slight alteration, such as having a bowl blocking the trajectory as illustrated in \figref{fig:baxterImitation}. 
Here we assume that the robot can only be able to receive these constraints during task execution phase (testing phase), and these constrains are not presented during the training phase. In this work, we propose an optimization based framework to adapt trained movements for novel environments. The first goal of our system is to generate adapted trajectories, as shown in \figref{fig:baxterAdaptation}, that can: 1) follow the demonstrated trajectories for the purpose of preserving movement patterns, and 2) fulfill novel constraints perceived from the environment during testing phase. 

Moreover, novel environment constraints perceived during testing phase could be more complicated than just obstacles. Following the example mentioned before, this time let's consider a situation where the target bottle is leaking. Ideally an intelligent robot that understands the situation should avoid moving the bottle over the bowl, but follows the movement path \textit{around} it. Even though we could adjust the objective function during optimization for movement adaptation, what if in another scenario the robot is asked to transfer a knife while avoiding obstacles \textit{above} them to prevent potential scratches? Such constraints are not only associated with the task context, i.e, leaking bottle or knife as the manipulated object, but also associated with user's preference, i.e, avoiding the bowl with a certain manner. Therefore, a human-in-the-loop on-line adaptation system is necessary to generate manipulation trajectories for different preferences. In the optimization framework presented in this paper for movement adaptation, we first treat the reward weights as the adjustable parameters to alter the quality of the trajectory. Then based on user feedback, the framework learns the preferred behavior, that fulfills constraints, by updating the reward weights. Therefore, the learned behavior can be generalized towards different situations where similar constraints are encountered. As illustrated in \figref{fig:baxterLearnAdaptation}, after a few iterations of on-line learning, the robot is able to generate an adapted trajectory according to the learned preferences.

This paper proposes an approach for interactively learning movement adaptation for manipulation tasks. \figref{fig:Block_diagram_adapt} illustrates the proposed system. The main contributions of this work are: 1) A system for robot to generalize movement learned from demonstrations to fulfill constraints perceived from novel environment. It is able to adapt trajectories for various situations according to user preferences; 2) An approach for robot learning to adapt trajectories by updating reward weights based on users' feedback. The user thus can co-actively train the robot in-the-loop by demonstrating desired trajectories; 3) An implementation of the optimization schema to adapt transferring skill considering obstacles and different manners. We validate the implementation on a humanoid platform (Baxter) and the experimental results support our claims.

\section{Related Work} 

Various approaches have been proposed to enable robot learning manipulation movements. Among them, imitation learning~\cite{Asfour06} focuses on mimicking human demonstrations, while learning from demonstration (LfD) techniques~\cite{Pastor09} are applicable. However, with these approaches, the robot could only reproduce learned movement in a similar environment.
To deal with novel environments, extended approaches~\cite{kober2008learning} augmented the trajectory generation with additional cost terms or different objective function as a criterion of trajectories' quality. The criterion is based on human experts' prior knowledge about the task or environment before execution phase. Then, the motion is generalized with these predefined constraints in similar situations. These approaches do not consider various user preferences. Here, we present another layer of exploration and learning to adapt the trained movement considering novel environment constraints, such as observed obstacles and task preferences.

Approaches~\cite{gams2009line} for encoding trajectory as motion primitives have been proposed for various forms of generalization and modulation, such as Gaussian mixture regression and Gaussian mixture models~\cite{guenter2007reinforcement, Calinon10}. In~\cite{khansari2010imitation}, a mixture model was used to estimate the entire movement skill from several sample trajectories. Another school of approaches derive from Hidden Markov models~\cite{inamura2004embodied}.
One popular representation to encode motion from demonstrated trajectories is Dynamic Movement Primitives (DMPs), as introduced in~\cite{Ijspeert13}. It consists of differential equations with well-defined attractor properties and a non-linear learnable component that allows modeling of almost arbitrarily complex motion. Recently, Probabilistic Movement Primitives (ProMPs)~\cite{paraschos2013probabilistic} was proposed as an alternative representation in probabilistic formulation. It learns a trajectory distribution from multiple demonstrations and modulates the movement by conditioning on desired target states. Incorporating the variance of demonstrations, ProMPs approach handles noise from different demonstrations and provides increased flexibility for reproducing movement. However, all these approaches hardly deal with novel environments such as involving different obstacles. In our work, we first train our robot using ProMPs and then generalize these trained motion primitives to newly introduced environment constraints.



In order to enable MPs to adapt to novel environments with obstacles~\cite{park2008movement, hoffmann2009biologically}, Kober et al.~\cite{kober2008learning} proposed an augmented version of DMPs which incorporates perceptual coupling to an external variable. They firstly learned the initial dynamic models by standard imitation learning and subsequently used a reinforcement learning method for self-improvement. Ghalamzan et al.~\cite{ghalamzan2015incremental} proposed a three-tiered approach for robot learning from demonstrations that can generalize noisy task demonstrations to a new target state and to an environment with obstacles. They encoded the nominal path generated from a Gaussian Mixture Model with DMPs and generated trajectory for a new target state. Then they adapted the DMP-generated trajectory to avoid obstacles by formulating an optimal control problem regarding the reward function learned from demonstrations by inverse optimal control. This approach allows an non-expert user to teach a robot the desired response to different objects but requires offline training in the environment involving those obstacles for learning the reward function. However, in real world scenarios, the human users often have different preferences for trajectories generation according to various environments and tasks, while it is extremely challenging for them to provide the optimal trajectories in every situation. Instead, in our approach, the human users can interactively provide sub-optimal suggestions on how to improve the trajectory and the robot learns the preference for different constraints, and also incorporate it in generating more applicable trajectories.

User preferences over robot's trajectories have been studied in the field of human robot interaction (HRI). Sisbot et al.~\cite{sisbot2007spatial} proposed to model user specified preferences as constraints on the distance of the robot from the user, the visibility of the robot and the user’s arm comfort. Then a path planner fulfilling such user preferences is provided. Ashesh Jain et al.~\cite{jain2015learning} proposed a co-active learning method to learn user preferences over generated trajectories for manipulation tasks by iteratively taking user sub-optimal feedback, thereafter the optimal trajectory was selected based on the learned reward function. In our work, we adopt the co-active learning paradigm and further propose a reward formulation to model user preferences over constraints for movement generation. Then we integrate it with movement adaptation through optimization based planning.

\section{Co-active Learning for Movement Generalization}
For the problem of robot learning from demonstrations~\cite{Pastor09}, a common practice is to offline learn the skills by encoding the trajectories with movement patterns such as DMPs~\cite{mao2014learning}. They can be then, during the testing phase, used to generalize the movement to novel situations with slight alterations, such as different initial and target states. Nevertheless, this generalization capability does not apply to novel environments with different obstacles or to a new task contexts with a variety of manipulated objects. In this paper, we propose a complementary framework for generalizing movement skills, which are offline learned from demonstrations, to novel situations, and in addition incorporate on-line learning preferences of how to generalize from human's feedback co-actively.

While facing a novel situation, the robot is given a manipulation task context $\bm{x}_c$ that describes the environment, the objects and any other task-related information. It could compute an imitation movement trajectory $\bm{y}_{D}$ by generalizing offline learned skills to new initial and target states. Such a trajectory can be executed if the new environment does not have obstacles and there is no other constraints inherited from the task. 

To further generalize learned movement skills to more challenging situations, the robot has to generate an adapted trajectory $\bm{y}$ based on the task contexts $\bm{x}_c$ and the computed imitation trajectory $\bm{y}_D$. Here we use a reward function $f^*(\bm{y}, \bm{x}_c, \bm{y}_D)$ to reflect how much reward the adapted trajectory $\bm{y}$ can achieve for different contexts. Therefore, we can adapt the movement by solving an optimal control problem which outputs an adapted trajectory by maximizing the reward function $f^*$. The reward function consists of a Imitation Reward $f_D$ describing the tendency to follow the imitation trajectory $\bm{y}_D$, a Control Reward $f_C$ describing the smoothness of executing the adapted trajectory $\bm{y}$ and a Response Reward $f_E$ describing the expected response given the environment. Although this reward function can be recovered from demonstrations by Inverse Optimal Control, as~\cite{ghalamzan2015incremental} suggests, it assumes that demonstrations are from experts, which bears an oracle reward function. In fact, it is common for non-expert users to provide non-optimal trajectories in practice. Also,~\cite{ghalamzan2015incremental} requires the manipulated objects or obstacles existing during these demonstrations and is hard to update the learned reward function online when the robot is facing situations that involve new objects. 
To learn the reward function which controls how the robot adapts trajectories under new contexts, we applied a co-active learning technique~\cite{jain2015learning} in which the user only corrects the robot by providing an improved trajectory $\bm{\bar{y}}$ and then the robot updates the parameter $\bm{w}$ of $f(\cdot; \bm{w})$ based on user's feedback. It is worth to note that this feedback only indicates $f^*(\bm{\bar{y}}, \bm{x}_c, \bm{y}_D) > f^*(\bm{y}, \bm{x}_c, \bm{y}_D)$, and $\bm{\bar{y}}$ may be non-optimal trajectories. With iterations of improvement, the robot could learn a function that approximates the oracle $f^*(\cdot)$ tightly.

\section{Our System}
Overall, after the robot has offline learned the movement skill from demonstrations, when  facing a different task context $\bm{x}_c$ in a novel environment, the testing phase includes three stages: 1) Movement Imitation, which computes an imitation trajectory $\bm{y}_{D}$ by generalizing demonstrated movement to new initial and target states; 2) Movement Adaptation, which generates an adapted trajectory $\bm{y}$ under new task and environment contexts by maximizing the given reward function; 3) Rewards Learning, which updates the parameters of estimated reward function according to user's feedback through co-active learning.  \figref{fig:Block_diagram_adapt} demonstrates our proposed framework. In the following sections, we entail and formulate each stage.

\subsection{Movement Imitation}
At the beginning, our system offline learns movement skill in an environment without obstacle or other constraints. In this work, we adopt the Probabilistic Movement Primitives (ProMPs)~\cite{paraschos2013probabilistic} for offline learning and movement imitation. It obtains a distribution over trajectories from multiple demonstrations, which captures the variations, and can be easily generalized to new initial and target states while imitating the movement. 

To be specific, we consider that a robot's end-effector has $d$ degrees of freedom (DOF) along with its arm, with its state denoted as $\bm{y}(t) = [y_1(t), \dots, y_d(t)]^T$. The trajectory of the robot's end effector is represented as a sequence $\bm{\mathcal{T}}=\left\{\bm{y}(t)\right\}_{t=0,\dots,T}$. We model each dimension $i$ of $\bm{y}(t)$ using linear regression with $n$ Gaussian time-dependent basis functions $\psi$ and a $n$-dimensional weight vectors $w_i$ as 
\begin{equation}
  y_i(t) = \psi(t)^Tw_i + \epsilon_y,
\end{equation}
where $\epsilon_y \sim \mathcal{N}(0,\sigma_y^2)$ denotes zero-mean i.i.d. Gaussian noise. With the underlying weight vectors $\bm{w}=[w_1^T, \dots, w_d^T]^T$, the probability of observing a trajectory $\bm{\mathcal{T}}$ can be given by
\begin{equation}
\label{eq:distoftrajectory}
p(\bm{\mathcal{T}}|\bm{w}) = \prod_tp(\bm{y}(t)|\bm{w}) = \prod_t\mathcal{N}(\bm{y}(t)|\bm{\Psi}(t)^T\bm{w},\bm{\Sigma_y}) 
\end{equation} 
where $\bm{\Psi}(t)=\text{diag}(\overbrace{\psi(t),\dots,\psi(t)}^{d})$ and $\bm{\Sigma_y} = \sigma_y^2\bm{I}_{d\times d}$. 

\subsubsection{Learning from Demonstrations} For each demonstration, the trajectory can be easily represented by a weight vector $\bm{w}$ which has fewer dimensions than the number of time steps. To capture trajectory variations from multiple demonstrations of the movement, a Gaussian distribution $p(\bm{w};\bm{\theta}) = \mathcal{N}(\bm{w}|\bm{\mu_w},\bm{\Sigma_w})$ over the weights $w$ is estimated. Therefore, the distribution of the trajectory $p(\bm{\mathcal{T}}|\bm{w})$ can be represented as
\begin{align}
\label{eq:distofw}
p(\bm{\mathcal{T}}; \bm{\theta}) &= \int p(\bm{\mathcal{T}}|\bm{w})p(\bm{w};\bm{\theta})d\bm{w} \\
  &= \prod_t\mathcal{N}(\bm{y}(t)|\bm{\Psi}(t)^T\bm{\mu_w}, \bm{\Psi}(t)^T\bm{\Sigma_w}\bm{\Psi}(t)^T+\bm{\Sigma_y})
\end{align}
We can then estimate the parameters $\bm{\theta}=\{\bm{\mu_w},\bm{\Sigma_w}\}$ by using maximum likelihood estimation as suggested in~\cite{paraschos2013probabilistic}.

\subsubsection{Trajectory Generation}
\label{subsec:trajectory}
In novel situations, the trajectory could be modulated by conditioning with different observed states. By adding an observation vector of $\bm{Y^*} = [\bm{y}_0^{*T}, \bm{y}_T^{*T}]^T$ indicating desired initial state $\bm{y}_0^*$ and target state $\bm{y}_T^*$ with the accuracy $\bm{\Sigma_y}^*$, we could apply Bayes theorem and represent conditional distribution for $\bm{w}$ as
\begin{equation}
\begin{array}{ll}
&p(\bm{w}|\bm{Y^*}) = \mathcal{N}(\bm{w}|\bm{\mu_w}', \bm{\Sigma_w}') \propto \mathcal{N}\left(\bm{Y}^*|\bm{\Psi}^{*T}\bm{w}, \bm{\Sigma_Y}^*\right)p(\bm{w}) \\
&\bm{\mu_w}' = \bm{\mu_w} + \bm{\Sigma_w}\bm{\Psi}^*\left(\bm{\Sigma_Y}^*+\bm{\Psi}^{*T}\bm{\Sigma_w}\bm{\Psi}^*\right)^{-1}\left(\bm{Y}^* -\bm{\Psi}^{*T}\bm{\mu_w}\right) \\
&\bm{\Sigma_w}' = \bm{\Sigma_w} - \bm{\Sigma_w}\bm{\Psi}^{*}\left(\bm{\Sigma_Y}^*+\bm{\Psi}^{*T}\bm{\Sigma_w}\bm{\Psi}^*\right)^{-1}\bm{\Psi}^{*T}\bm{\Sigma_w}
\end{array} 
\end{equation}
where $\bm{\Psi}^{*}=[\bm{\Psi}(0), \bm{\Psi}(T)]$ and $\bm{\Sigma_Y}^*=\text{diag}(\bm{\Sigma_y}^*, \bm{\Sigma_y}^*)$ are augmented for observation vector $\bm{Y}^*$. 


With a conditional distribution of $\bm{w}$, we could generate conditional trajectory distribution and easily evaluate the mean $\bm{y}_D$ and the variance $\bm{\Sigma}_D$ of the trajectory $\bm{\mathcal{T}}$ for any time point $t$ according to Eq.(~\ref{eq:distoftrajectory}) and Eq.(~\ref{eq:distofw}). Therefore, the mean trajectory ${\bm{y}_D(t)}$ can be used as the imitation trajectory in movement adaptation and the variance $\bm{\Sigma}_D(t)$ can be used to indicate which parts or dimensions of the trajectory are more flexible to adapt. The larger variance reflects higher variations in demonstrations. It means more flexibility to modify the corresponding part of the trajectory. 

It is worth to mention that, although we adopt ProMPs for movement imitation in this work, the proposed Movement Adaptation framework can be integrated with any other movement imitation learning techniques.

\subsection{Movement Adaptation}
As mentioned before, if the environment of a new situation is exactly the same as the one during demonstration when ProMPs are learned, e.g, no obstacle, safety constraints or other new considerations, the robot can perform movement optimally by directly following the imitation trajectory $\bm{y}_D \in \R^d$ in discrete time generated by learned ProMPs.

In this work, we want to have a system that can adapt to an environment with novel constraints. 
Thus, we model the movement adaptation as an optimal control problem with fixed time horizon $T$ in discrete time. The output of the adaptation system is a new trajectory $\bm{y} \in \R^d$ in discrete time. The input consists of the task context $\bm{x}_c$ that describes the environment, the objects and any other task-related information which are obtained from the perception module, the imitation trajectory $\bm{y}_D$ which is generated from learned ProMPs, and the reward function $f(\bm{y}, \bm{x}_c, \bm{y}_D)$ which represents the reward of the adapted trajectory $\bm{y}$ corresponding to the new situation. 

\subsubsection{Optimization with Constraints} Let's consider that the perception module detects $N_{obj}$ objects in the environment, which may be obstacles during the manipulation. Each object is abstracted as a sphere in the space represented by its center location and semi-diameter $\{\bm{O}_k, d_k\},k=1,\dots,N_{obj}$. Assuming the reward function can be modeled as accumulated sum of rewards from each state $\bm{y}(t)$ at time step $t$: 
\begin{equation}
\label{eq:discreterewards}
 f(\bm{y}, \bm{x}_c,\bm{y}_D) = \sum_{t=0}^{T}f_t(\bm{y}(t), \bm{x}_c,\bm{y}_D).
\end{equation}
Because we are only modulating the trajectory, we can model the adaptation system as linear dynamics with the control signal $\bm{a} \in \R^m$, as it does not involve real physical dynamics. According to the embodiment of robotic end-effector based on its design, we could compute the end-effector's position in spatial space $\bm{E}(\bm{y})$ following the kinematics modeling~\cite{ju2014kinematics}. Then, considering obstacles avoidance in spatial space, the target optimal policy $\pi^* = \{\bm{a}(t)^*\}_{t=0,\dots,T-1}$ could be defined from Eq. (\ref{eq:optimal}) with constraints.
\begin{eqnarray}
\label{eq:optimal}
&\pi^*=\argmax{\pi} \sum_{t=0}^Tf_t(\bm{y}(t), \bm{x}_c, \bm{y}_D)  \\ 
\text{subj. to} & \forall t = 0,\cdots, T-1\\
& \bm{z}(t+1) = \bm{A}\bm{z}(t) + \bm{B}\bm{a}(t) \\
&  \bm{y}(t) = \bm{C}\bm{z}(t) \\
\label{eq:limits}
& \bm{U} \geq \bm{y}(t) \geq \bm{L} \\
\label{eq:obst}
& \lVert\bm{E}(\bm{y}(t))-\bm{O}_k\rVert^2 \geq d_k^2, \quad\forall k=1,\cdots,N_{obj} \\
\label{eq:final}
& \bm{y}(T) = \bm{y}_D(T),
\end{eqnarray}
where $\bm{A}, \bm{B}, \bm{C}$ are system matrices, Eq.(~\ref{eq:final}) constrains the final position of the adapted trajectory, Eq.(~\ref{eq:limits}) constrains the trajectory within feasible limits, and Eq.~\ref{eq:obst} ensures the adapted trajectory can avoid obstacles safely by keeping a minimum distance $d_{k}$ between the robot's end-effector and any object.

\subsubsection{Model Predictive Control}
 In order to find an optimal solution of such a system with continuous state and action spaces, we adopt Model Predictive Control which computes the optimal actions in a finite prediction horizon. Therefore, by considering a prediction time horizon $T_p$, the optimal action $\bm{a}(i)^*$, at time step $i = 0,\dots, T-1$, can be solved by:
\begin{equation}
\label{eq:mpc}
\begin{array}{rc}
\multicolumn{2}{c}{\umax{(\bm{a}(i), \cdots, \bm{a}(i+T_p-1))} \sum_{t=i+1}^{i+T_p} f_t(\bm{y}(t), \bm{x}_c, \bm{y}_D)} \\ 
\rule{0ex}{1em}\text{subj. to} &  \forall t = i, \cdots, i+T_p-1\\
\rule{0ex}{1em}&\bm{z}(t+1) = \bm{A}\bm{z}(t) + \bm{B}\bm{a}(t) \\
\rule{0ex}{1em}&  \bm{y}(t) = \bm{C}\bm{z}(t) \\
\rule{0ex}{1em}& \bm{U} \geq \bm{y}(t) \geq \bm{L} \\
\rule{0ex}{1em}&\lVert\bm{E}(\bm{y}(t))-\bm{O}_k\rVert^2 \geq d_k^2, \quad\forall k = 1,\cdots,N_{obj} \\
\rule{0ex}{1em}& \bm{y}(T) = \bm{y}_D(T) .
\end{array}
\end{equation}
At each step $i$, the optimal actions $\{\bm{a}(i)^*, \cdots, \bm{a}(i+T_p-1)^*\}$ for $T_p$ decision steps in future are computed but only the action for current step $\bm{a}(i)^*$ is performed. Therefore, it can deal with changing environments as these changes could be considered in the next decision steps.  

\subsubsection{Reward Function} In order to adapt robot movements to perform well in novel situations, considering only hard constraints such as obstacle avoidance, Eq.(~\ref{eq:obst}), does not suffice. Thus, our framework further models a reward function $f(\bm{y}, \bm{x}_c, \bm{y}_D)$ that reflects the amount of rewards that an adapted trajectory $\bm{y}$ can gain within the context $\bm{x}_c$ and $\bm{y}_D$.  As the reward function $f(\bm{y})$ is assumed temporally discrete in Eq.(~\ref{eq:discreterewards}), we model the reward function $f_t(\bm{y}(t))$ at  $t$ by three parts:
\begin{equation}
\label{eq:rewards}
f_t(\bm{y}(t);\bm{w}) = f_{D,t}(\bm{y}(t);\bm{w}_D) + f_{C,t}(\bm{y}(t);\bm{w}_C) + f_{E,t}(\bm{y}(t);\bm{w}_E),
\end{equation}
where the Imitation Reward $f_D$ models the tendency to follow the imitation trajectory $\bm{y}_D$, the Control Reward $f_C$ models the smoothness of executing the adapted trajectory $\bm{y}$ and the Response Reward $f_E$ characterize the expected response to the environment. Meanwhile, $\bm{w} = [\bm{w}_D^T,\bm{w}_C^T,\bm{w}_E^T]^T$ are parameters that affect the behavior of the movement adaptation. We describe each reward function in detail as follows.

\paragraph{Imitation Reward}
Imitation Reward characterizes how well the adapted trajectory can imitate the demonstrations through the distance between points on $\bm{y}$ and $\bm{y}_D$. 
Recall that we have the variance $\bm{\Sigma}_D(t)$ of the imitation trajectory $\bm{y}_D$ by Movement Imitation~\ref{subsec:trajectory}, which indicates how flexible we could adapt the trajectory. Considering $\bm{\Sigma}_D(t) = \text{diag}(\sigma^2_1(t),\dots, \sigma^2_d(t))$ to be diagonal for the sake of simplicity, we model the Imitation Reward by the weighted distance:
\begin{align}
  &f_{D,t}(\bm{y}(t);\bm{w}_D) = -(\bm{y}(t) - \bm{y}_D(t))^T\bm{V}(t)(\bm{y}(t) - \bm{y}_D(t)) \\
  &\bm{V}(t) = \text{diag}(\bm{w}_D)\text{diag}(e^{-\sigma^2_1(t)}, \dots, e^{-\sigma^2_d(t)}),
\end{align}
where $\bm{V}(t)$ is a weight matrix consisting of parameters $\bm{w}_D$ and $\{e^{-\sigma^2_i(t)}\}$ in which the variances learned from demonstrations $\bm{\Sigma}_D(t)$ are modeled to affect adaptation rewards.

\paragraph{Control Reward}
Control Reward $f_C$ characterizes the smoothness of executing the adapted trajectory $\bm{y}$ through the following formulation:
\begin{align}
  f_{C,t}(\bm{y}(t);\bm{w}_C) = -\bm{w}_C\lVert(\bm{y}(t) - \bm{y}(t-1))\rVert^2,
\end{align}
where $\bm{w}_C$ is the parameter to weigh this reward.

\paragraph{Response Reward}
Response reward $f_E$ describes the expected response to the environment such as safety considerations for obstacles and objects under manipulation. Here we give intuitive examples for Response Reward. Although we can ensure minimum distance to avoid obstacles using Eq.(~\ref{eq:mpc}), as human users we still expect the robot to transfer a cup full of water \textit{around} a laptop instead of \textit{above} it, in case of spilling. Another example is that the user would prefer the robot manipulating sharp objects, such as a knife, to keep a relatively larger distance from the human for safety consideration. These examples indicate that we would have preferences towards how the robot avoids obstacles. Moreover, for safety consideration, we also prefer the robot to transfer a fragile object closer to the table top to maintain a safety margin. All the above preferences are specific to objects under manipulation and the exact environment. Thus, we set the Response Reward to ensure that the better the adapted trajectory fulfills these preferences, the higher the reward is. 

\begin{figure}[thpb]
\centering
  \includegraphics[width=3in]{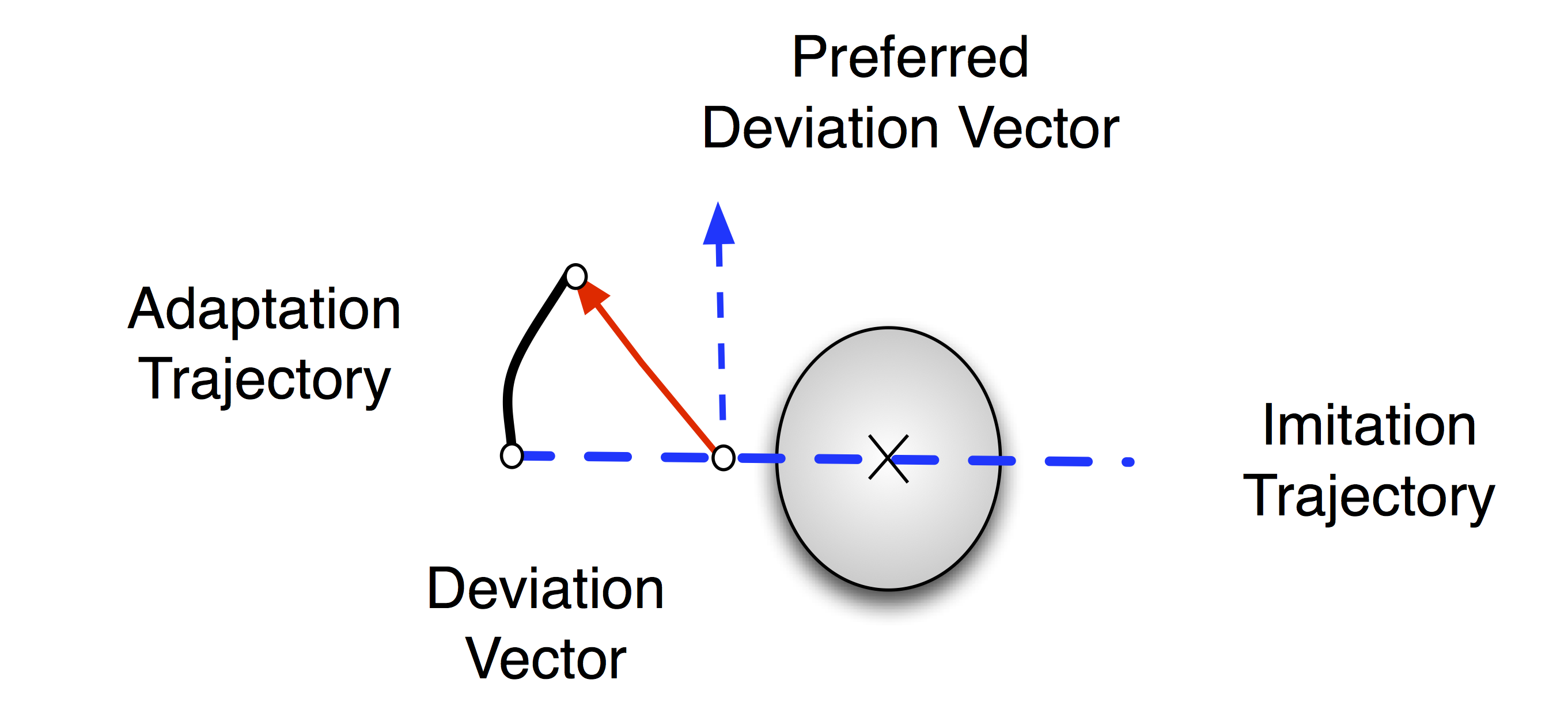} 
  \caption{Illustration of deviation vector feature: vector from original imitation trajectory to an adapted one.}
  \label{fig:deviation}
  \vspace{-10pt}
\end{figure}

To formally represent the Response Reward, let us consider a scenario with $N_{obj}$ obstacles on the table. The leftmost and rightmost locations of the table are $\bm{B}_1, \bm{B}_2$ and the table surface is  $\bm{S}$, we then can formulate Response Rewards as follows:
\begin{align}
    &f_{E,t}(\bm{y}(t);\bm{w}_E) = -\left(\sum_{k=1}^{N_{obj}}\bm{w}_{O,k}^T\bm{\phi}_{O,k}+w_{B}\phi_B + w_S\phi_S\right)\\
    &\begin{array}{rcl}
      \bm{\phi}_{O,k}^T &=& \left[-\lVert\bm{E}(\bm{y}(t))-\bm{O}_k\rVert, (\bm{E}(\bm{y}_D(t))-\bm{E}(\bm{y}(t)))^T\right] \\
      &\cdot&\text{exp}\left(-\frac{\lVert\bm{E}(\bm{y}(t))-\bm{O}_k\rVert^2}{d_{k}}\right)
    \end{array}\\
    &~~\phi_B = \sum_{i=1}^2\text{exp}\left(-\frac{\lVert\bm{E}(\bm{y}(t))-\bm{B}_i\rVert^2}{d_{min}}\right) \\
    &~~\phi_S = \lVert \bm{E}(\bm{y}(t)) - S\rVert^2,
\end{align}
where $\bm{\phi}_{O,k}$ represents the feature vector for preferences in avoiding obstacle $\bm{O}_k$, of which the first element denotes avoiding distance and the second element denotes the deviation vector as shown in \figref{fig:deviation}. The preferred deviation vector is given as reward weights and the inner product between two vectors indicates the rewards of deviation considering the given preference. The exponential decay function is applied so that the features are only effective when the robot's end-effector is close to the obstacles. $\phi_B$ and $\phi_S$ are features related to safety by considering boarders and surface of the table. $\bm{w}_E = [\bm{w}_{O,1}^T, \dots, \bm{w}_{O,N_{obj}}^T, w_B, w_S]^T$ are weights corresponding to the features respectively.

Given a set of parameters $\bm{w} = [\bm{w}_D^T,\bm{w}_C^T,\bm{w}_E^T]^T$, the MPC module generates an adapted trajectory by maximizing $f(\cdot;\bm{w})$. The robot could follow the adapted trajectory and execute the task facing the novel situation. However, the generated trajectory may not be satisfying enough from user's perspective, since the given or initialized parameters may not be accurate for modeling the rewards. To accommodate the issue, after the movement execution, our system allows the user to provide a better trajectory as feedback to update the parameters during the following Rewards Learning section.

\subsection{Rewards Learning}
In this section, we describe how our system learns the reward function. Assuming there is an oracle reward function $f^*(\bm{y}, \bm{x}_c, \bm{y}_D)$ that reflects exactly how much reward the adapted trajectory $\bm{y}$ can gain for each context. The goal of this module is to estimate such a reward function $f(\bm{y}, \bm{x}_c, \bm{y}_D; \bm{w})$, where $\bm{w}$ are the parameters to be learned, that approximate the oracle reward $f^*(\cdot)$ tightly. 

By rewriting Eq.(~\ref{eq:discreterewards}) and Eq.(~\ref{eq:rewards}) for the entire trajectory, we can have the reward function in a linear form represented by features and weights:
\begin{align}
&f(\bm{y}, \bm{x}_c,\bm{y}_D;\bm{w}) = \bm{w}_D^T\bm{\phi}_D + \bm{w}_C^T\bm{\phi}_C + \bm{w}_E^T\bm{\phi}_E \\
&\bm{\phi}_D = \left[\phi_{D,1},\dots,\phi_{D,d}\right]^T,\phi_{D,i} = -\sum_{t=0}^T\left(y_i(t)-y_{D,i}(t)\right)^2e^{-\sigma_i^2(t)} \\
&\bm{\phi}_{C} = - \sum_{t=1}^T\lVert(\bm{y}(t) - \bm{y}(t-1))\rVert^2 \\
&\bm{\phi}_E = -\sum_{t=0}^T\left[\bm{\phi}_{O,1}^T(\bm{y}(t)),\dots,\bm{\phi}_{O,N_{obj}}^T(\bm{y}(t)), \phi_B(\bm{y}(t)), \phi_S(\bm{y}(t))\right]^T
\end{align}
where $\bm{\phi}_D,\bm{\phi}_C,\bm{\phi}_E$ represent features of the entire trajectory corresponding to Imitation, Control and Response Rewards.

Since the user only provides a feedback trajectory $\bm{\bar{y}}$ and the system can not directly observe the reward function, we apply the co-active learning technique~\cite{jain2015learning} in which the robot iteratively updates the parameter $\bm{w}$ of $f(\cdot; \bm{w})$ based on user's feedback. Note that this feedback only needs to indicate $f^*(\bm{\bar{y}}, \bm{x}_c, \bm{y}_D) > f^*(\bm{y}, \bm{x}_c, \bm{y}_D)$ and $\bm{\bar{y}}$ could be non-optimal trajectories. Algorithm~\ref{algo:learning} gives our learning algorithm for movement adaptation.

\begin{algorithm}
  {
   \caption{Rewards Learning for Movement Adaptation}
   \label{algo:learning}
   \begin{algorithmic}
    \myState{Initialize $\bm{w}^{(0)} = [\bm{w}_D^{(0)T}, \bm{w}_C^{(0)T}, \bm{w}_E^{(0)T}]^T$}
    \For{Iteration $i=0$ to $T_l$}
      \myState{Task Context and Environment Perception: $\bm{x}_c^{(i)}$ }
      \myState{Movement Imitation:}
      \myState{$\quad\bm{y}_D^{(i)}, \bm{\Sigma}_D^{(i)} \leftarrow p(\bm{\mathcal{T}}|\bm{x}_c^{(i)})$}
      \myState{Movement Adaptation:}
      \myState{$\quad\bm{\pi}^{*(i)} = \arg\text{max}_{\bm{\pi}} f(\bm{y}, \bm{x}_c^{(i)}, \bm{y}_D^{(i)}; \bm{w}^{(i)})$}
      \myState{$\quad\bm{y}^{(i)} \leftarrow \bm{\pi}^{*(i)}$}
      \myState{Movement Execution: $\bm{y}^{(i)}$}
      \If{User Provides Feedback: $\bm{\bar{y}}^{(i)}$}
        \myState{$\alpha^{(i)} = 1/\sqrt{i}$}
        \myState{$\bm{w}_D^{(i+1)} = \bm{w}_D^{(i)} + \alpha^{(i)}(\bm{\phi}_D(\bm{\bar{y}}^{(i)},\bm{y}_D^{(i)}) -\bm{\phi}_D(\bm{y}^{(i)},\bm{y}_D^{(i)}))$}
        \myState{$\bm{w}_C^{(i+1)} = \bm{w}_C^{(i)} + \alpha^{(i)}(\bm{\phi}_C(\bm{\bar{y}}^{(i)}) -\bm{\phi}_C(\bm{y}^{(i)}))$}
        \myState{$\bm{w}_E^{(i+1)} = \bm{w}_E^{(i)} + \alpha^{(i)}(\bm{\phi}_E(\bm{\bar{y}}^{(i)}, \bm{x}_c^{(i)}) -\bm{\phi}_E(\bm{y}^{(i)},\bm{x}_c^{(i)}))$}
        \myState{Weights Projection:}
        \myState{$\quad \bm{\bar{w}}^{(i+1)} = [\bm{w}_D^{(i+1)T}, \bm{w}_C^{(i+1)T}, \bm{w}_E^{(i+1)T}]^T$}
        \myState{$\quad \bm{w}^{(i+1)} = \arg\text{min}_{\bm{w}\in\bm{C}}\lVert\bm{w}-\bm{\bar{w}}^{(i+1)}\rVert^2$}
      \Else{$\quad\bm{w}^{(i+1)} =\bm{w}^{(i)}$}
      \EndIf
    \EndFor
   \end{algorithmic}
   }
\end{algorithm}
Note that $\alpha$ is a learning rate, which decays along iterations and $\bm{C}$ in the weights projection part is a bounded set to ensure updated parameters $\bm{w}$ are in feasible space. After iterations of improvements, the robot can learn an estimated reward function $f(\cdot; \bm{w}^*)$ that approximates the oracle reward function $f^*(\cdot)$ as proved in ~\cite{cina2016proving}. By maximizing the estimated reward function $f(\bm{y}, \bm{x}_c, \bm{y}_D;\bm{w}^*)$, the robot can generate an adapted trajectory $\bm{y}$ that maximizes the rewards facing situation $\bm{x}_c$ based on imitation trajectory $\bm{y}_D$.

\section{Experiments}
To validate the system described above, we design and conduct the following experiments on a Baxter humanoid platform. The Baxter robot is asked to do manipulation tasks such as cleaning on a table top, with the surface as $\bm{S}=(0,0,-0.1)$, the leftmost location as $\bm{B}_1=(0,0.8,0)$ and the rightmost location as $\bm{B}_2=(0,-0.8,0)$ in robot spatial space in meter. It needs to learn transferring the manipulated object between different locations while avoiding obstacles with desired manners. 

\begin{figure}[!ht]
  \subfigure[]{
    \centering
    \includegraphics[width=0.46\columnwidth]{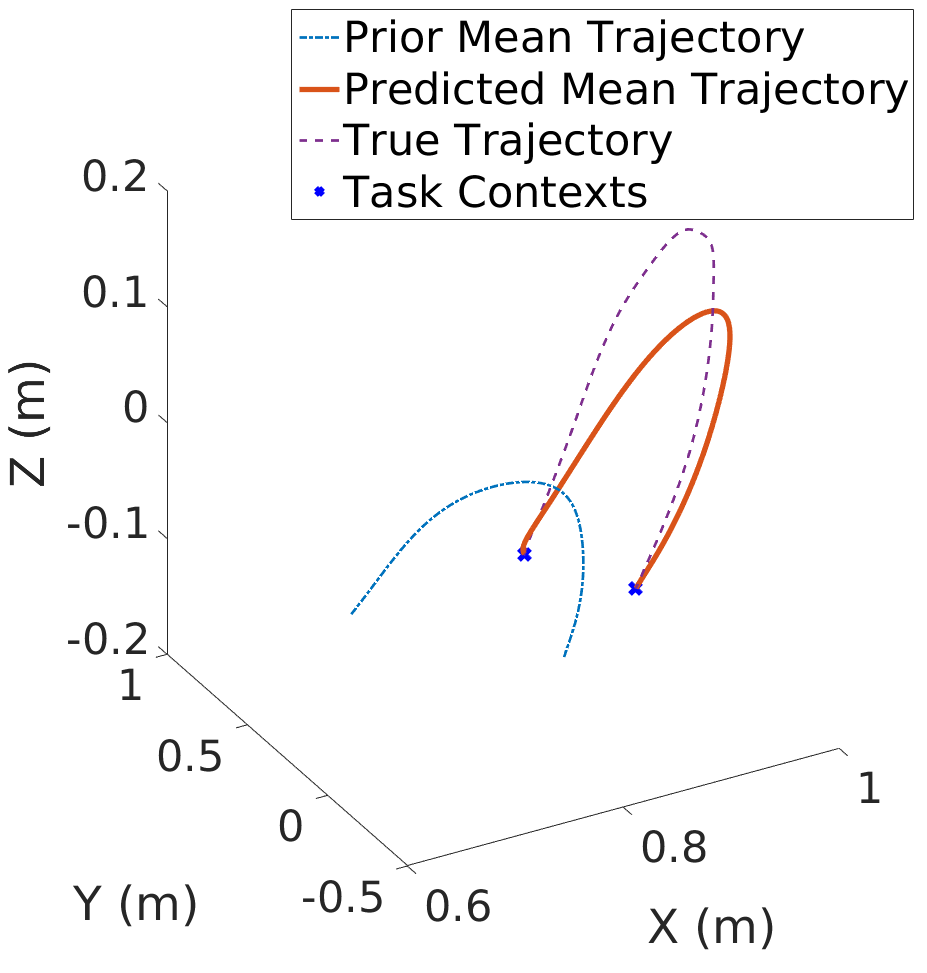} 
    \label{fig:imitation}
  }
  \subfigure[]{
    \centering
    \includegraphics[width=0.46\columnwidth]{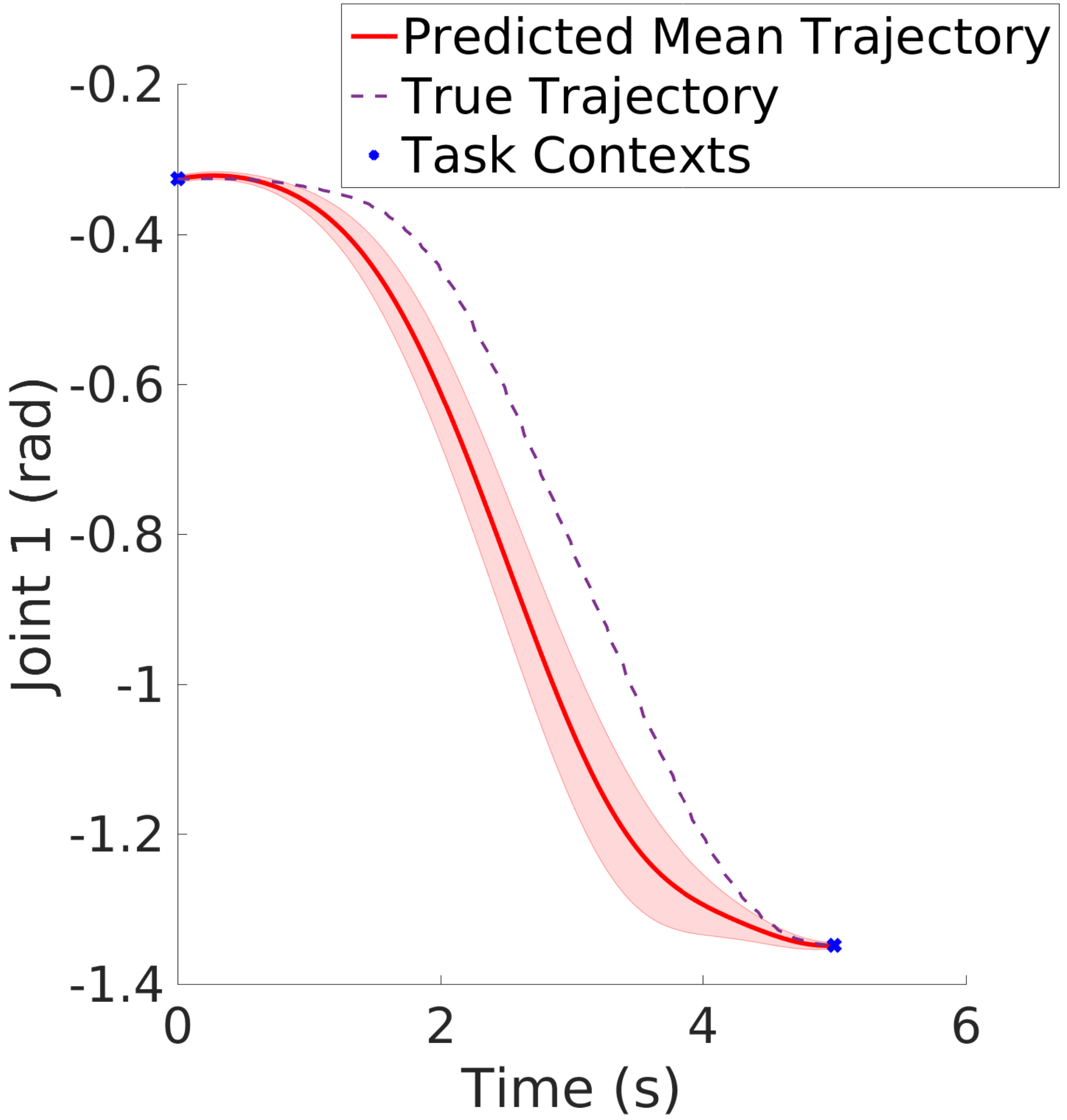} 
    \label{fig:imitationTraj1}
  }
  \caption{Movement Imitation with ProMPs for Transferring Task: (a) Imitation trajectory predicted based on prior movement and task contexts in spatial space; (b) Imitation trajectory for joint $s_0$ in joint space, shaded area indicating the predicted variance. }
  \vspace{-10pt}
\end{figure}

\begin{figure*}[!ht]
  \subfigure[]{
    \centering
    \includegraphics[height=2in]{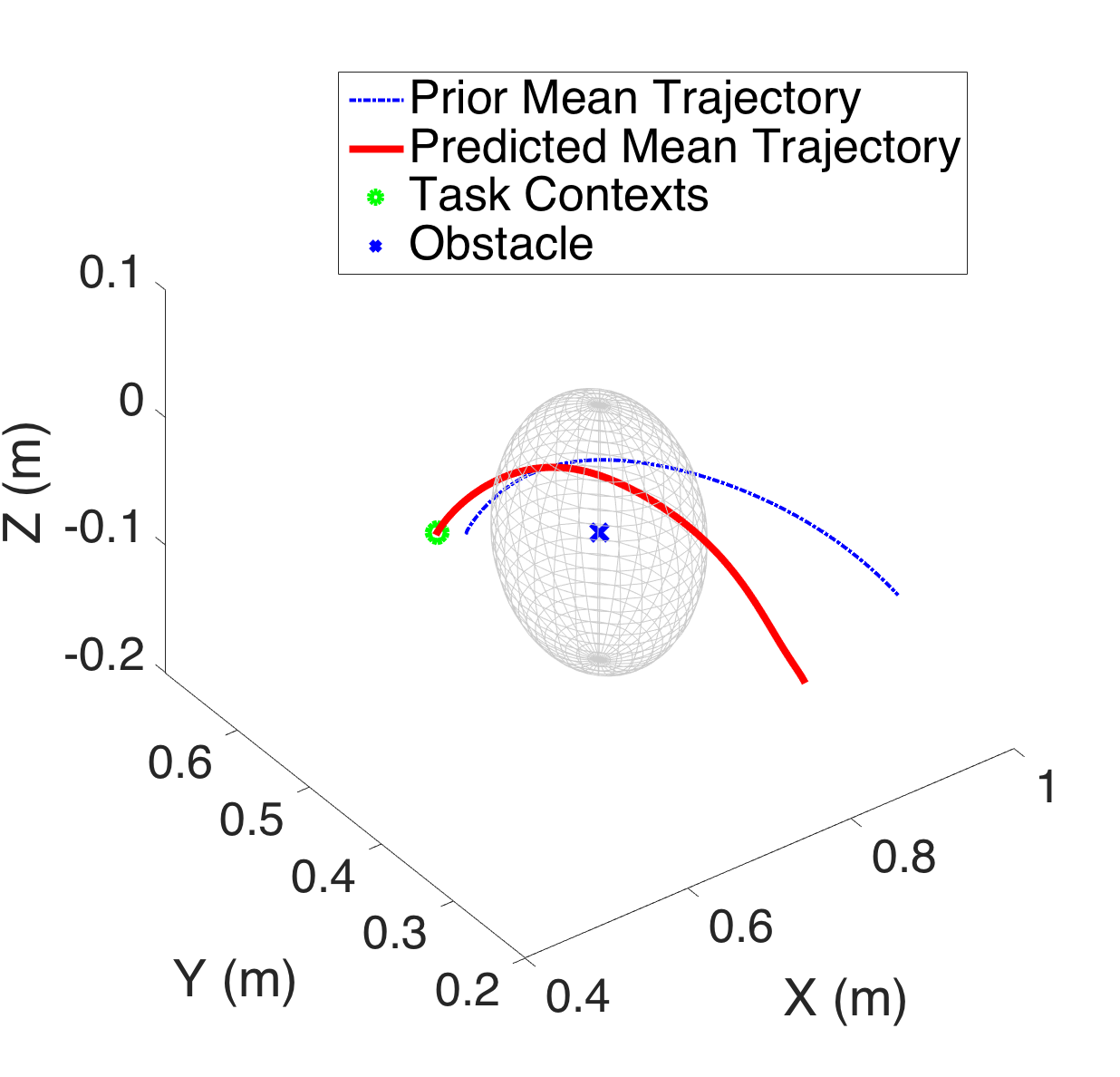} 
    \label{fig:bottleimitation}
  }
  \subfigure[]{
    \centering
    \includegraphics[height=2in]{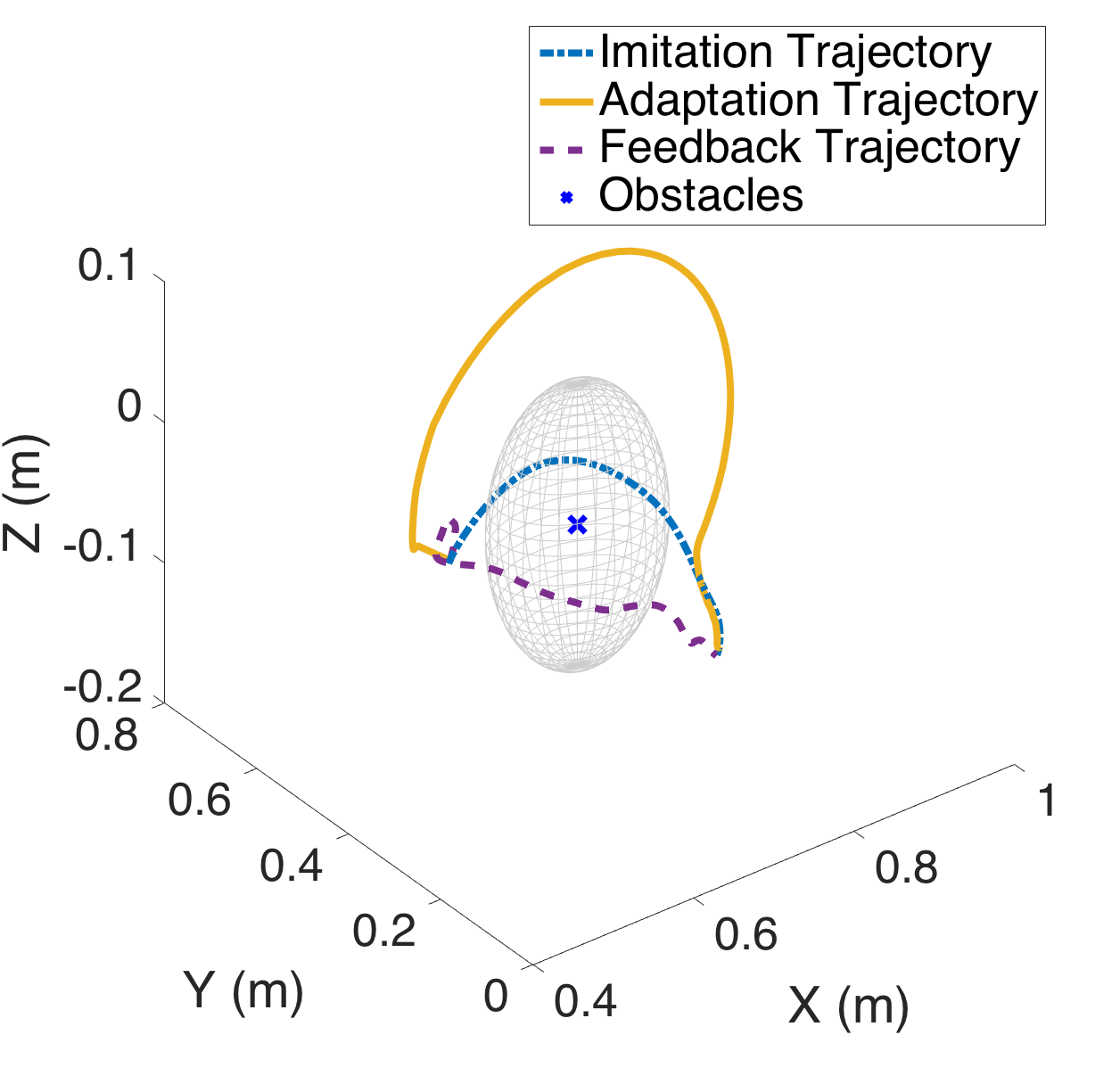} 
    \label{fig:bottleadaptbefore}
  }
  \subfigure[]{
    \centering
    \includegraphics[height=2in]{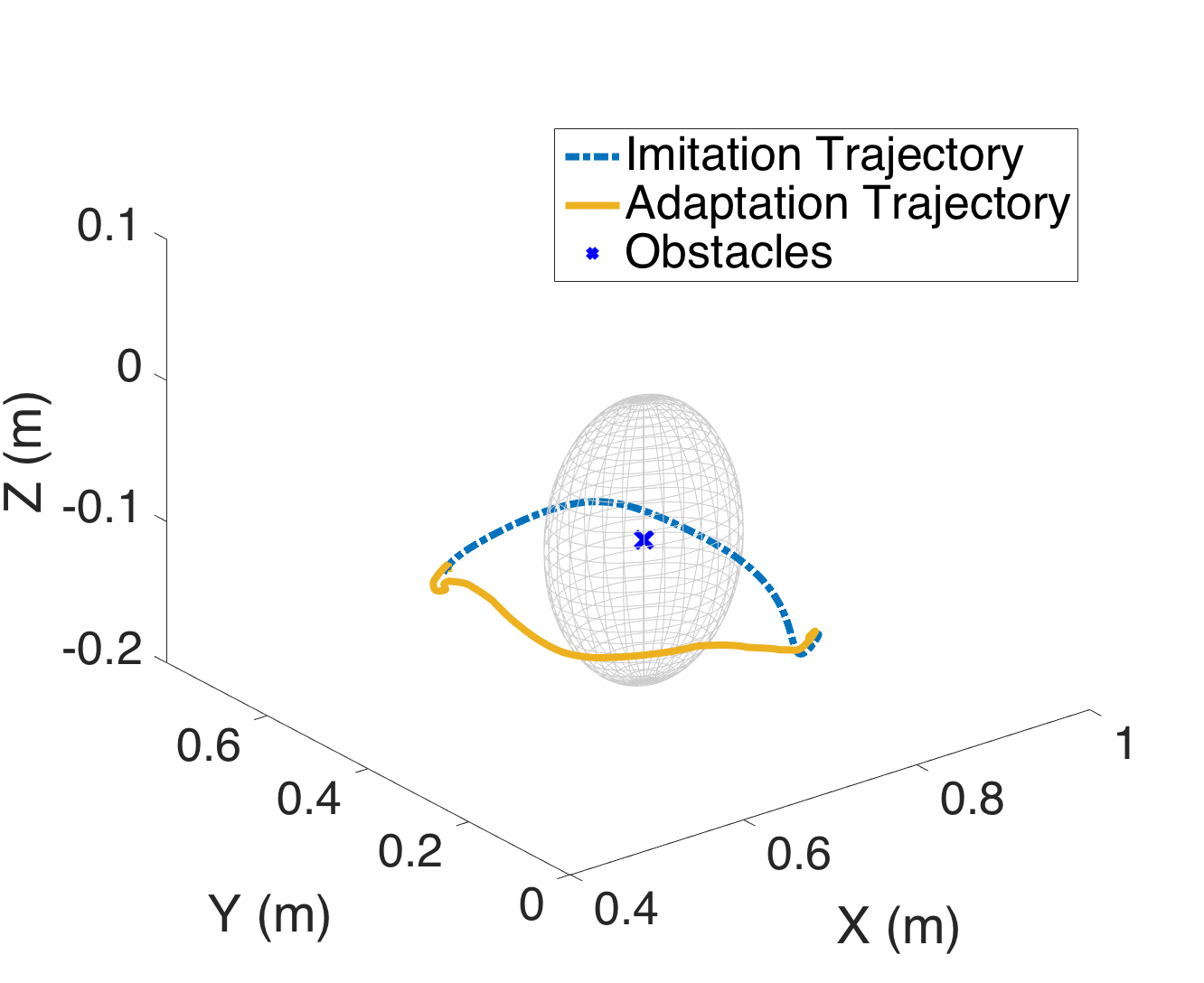} 
    \label{fig:bottleadaptafter}
  }
  \caption{Learning to Adapt Movement for Transferring a Leaking Bottle: (a) Movement Imitation, failed to avoid the obstacle; (b) Movement adaptation with initial weights, successfully avoided the obstacle by path above it but has a potential danger of spilling water, feedback trajectory is provided afterwards; (c) Movement adaptation for a different situation with new task contexts and obstacle locations, with updated weights after learning from feedback trajectory, successfully avoids the obstacle through a path around. Corresponding execution on the Baxter platform is given by \figref{fig:BaxterExecution}. }
  \vspace{-10pt}
\end{figure*}

During an off-line learning phase, the robot learns the movement skill from multiple kinethestic demonstrations with no obstacles on the table. During the online learning stage, a variety of obstacles are located randomly on the table and we assume the robot can obtain their locations from perception modules. The system learns iteratively to adapt the movement skill in novel situations such as \textit{with different manners avoiding} obstacles, at the same time \textit{follows the similar movement pattern} from off-line demonstrations.

\subsection{Movement Imitation}
In the first stage of the experiments, we have our robot learn off-line the movement skill from demonstrations. All trajectories are sampled discretely and normalized to $T=200$ steps for transferring movement in joint space, and the left arm of the Baxter has $d=7$ degree of freedom. The training trajectories are encoded by ProMPs with $n=10$ Gaussian basis functions so that the movement skill can be generalized to different initial and target states.

\figref{fig:imitation} shows an example of our generated imitation trajectory in spatial space for new task contexts using ProMPs. \figref{fig:imitationTraj1} shows the corresponding imitation trajectory of joint $s_0$ in joint space. The blue crosses here are desired new initial and target states, and the shaded area is the estimated variance for imitation trajectory, which reflects the variations of demonstrations. True trajectory here means a trajectory recorded from user demonstration in the testing scenario for comparison. It is not hard to see that the predicted mean of the imitation trajectory is well generalized to new initial and target states and follows the same movement pattern as the prior mean trajectory learned from demonstrations. Therefore, the robot can perform the task well by following this imitation trajectory if there is no obstacles or other safety constraints under new situations. 

\subsection{Learning Adaptation}
While facing a task of transferring a leaking bottle, the robot may find a bowl with food inside as an obstacle on the table where its center location $\bm{O}_1$ and minimum safety distance $d_1$ are assumed to be obtained through perception. 

For movement adaptation, we set the prediction horizon $T_p=11$ in model predictive control and select system matrices $\bm{A} = 0.9\cdot\bm{I},\bm{B}=\bm{C}=\bm{I}$ to make the system stable in the prediction window as suggested in~\cite{ghalamzan2015incremental}. The limits of joints could be found from the Baxter hardware specification. The minimum safety distance with table boarder is set as $d_{min}=0.1$. And the weights for reward function are initialized to be $\bm{w}_D=30\cdot\mathbf{1}, \bm{w}_C=10, \bm{w}_E=\mathbf{0}$. And then we apply the native Matlab Gradient-based optimization method \textit{fmincon} to solve the optimization at each time step.

\figref{fig:bottleimitation} shows the output from movement imitation for transferring the leaking bottle, which failed to avoid the obstacle even though the trajectory generalizes to a novel initial and target states. \figref{fig:bottleadaptbefore} shows the movement adaptation with initial weights. There is no preference specified in reward function about how to avoid obstacles or take safety considerations about boarders. Therefore, even though the adapted trajectory could avoid the obstacle successfully, it may be not an ideal trajectory.

\begin{figure}[!h]
  \vspace{-5pt}
  \subfigure[]{
    \centering
    \includegraphics[height=1.15in]{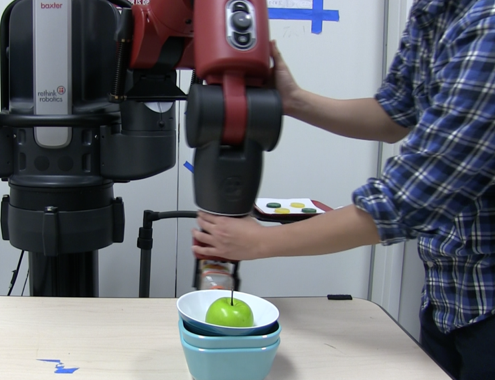} 
    \label{fig:feedback}
  }
  \subfigure[]{
    \centering
    \includegraphics[height=1.3in]{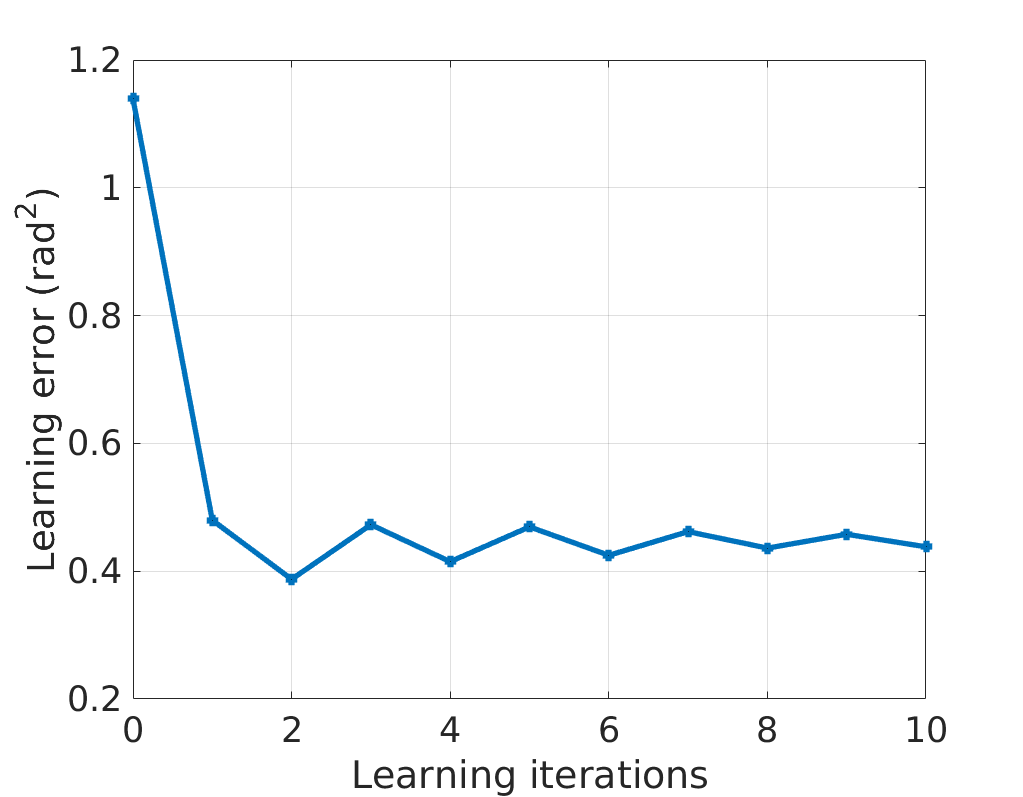} 
    \label{fig:learningcurve}
  }
  \caption{Rewards Learning from User Feedback for Transferring Leaking Bottle: (a) User feedback via kinethestic demonstration; (b) Learning curve for adaptation under the same feedback.}
  \vspace{-5pt}
\end{figure} 
To learn the user preference, we then provide feedback via kinethestic demonstration illustrated in \figref{fig:feedback} and the feedback trajectory is shown in \figref{fig:bottleadaptbefore} as dash line to indicate user preferences. Following Algo.~\ref{algo:learning}, the robot iteratively updates the rewards weights based on the user feedback. Weights are limited via projection in the feasible set $\bm{C}$ where $\bm{w}_D \in [1,100]^7, \bm{w}_C\in [1,100], \bm{w}_E\in[0,100]$ except that last two parameters in $\bm{w}_{O,k}$ indicating preferred deviation direction could be $[-100,100]$. To quantitatively validate the performance of our method in movement adaptation, we consider the metric of cumulative error between the adapted trajectory and the feedback trajectory $e(i) = \frac{1}{T}\sum_{t=0}^T\left(\bm{\bar{y}}^{(i)}(t)-\bm{y}^{(i)}(t)\right)^2$ as the learning error at iteration $i$. Since the metric is affected by different situations such as obstacles' locations, we consider the feedback trajectory as fixed and let the robot iteratively learn several times to see how it performs and record the ``learning curve'' under the same feedback. From \figref{fig:learningcurve}, we can see that the error decreases and converges after several iterations, and it only requires a few of iterations to achieve an adapted trajectory as desired preference according to the feedback. 

After learning, the robot uses the updated weights for movement adaptation in a different situation with novel initial/target states and the obstacles' locations. \figref{fig:bottleadaptafter} shows the adapted trajectory based on the updated weights after one iteration, where it successfully avoids the obstacle via the desired direction. 

\begin{figure}[!h]
  \subfigure[]{
    \centering
    \includegraphics[height=1.15in]{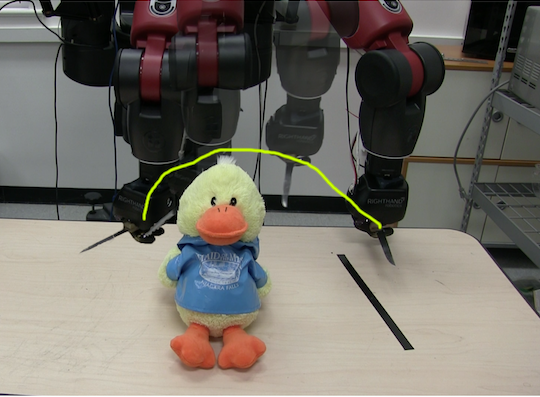} 
    \label{fig:baxterduckadaptation}
  }
  \subfigure[]{
    \centering
    \includegraphics[height=1.15in]{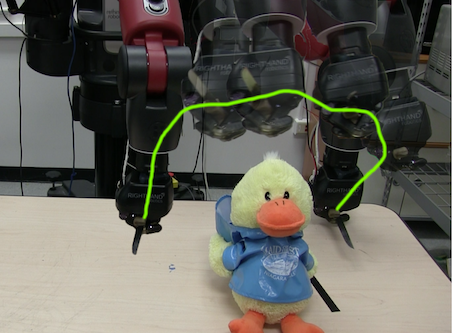} 
    \label{fig:baxterduckimproved}
  }
  \subfigure[]{
    \centering
    \includegraphics[height=1.45in]{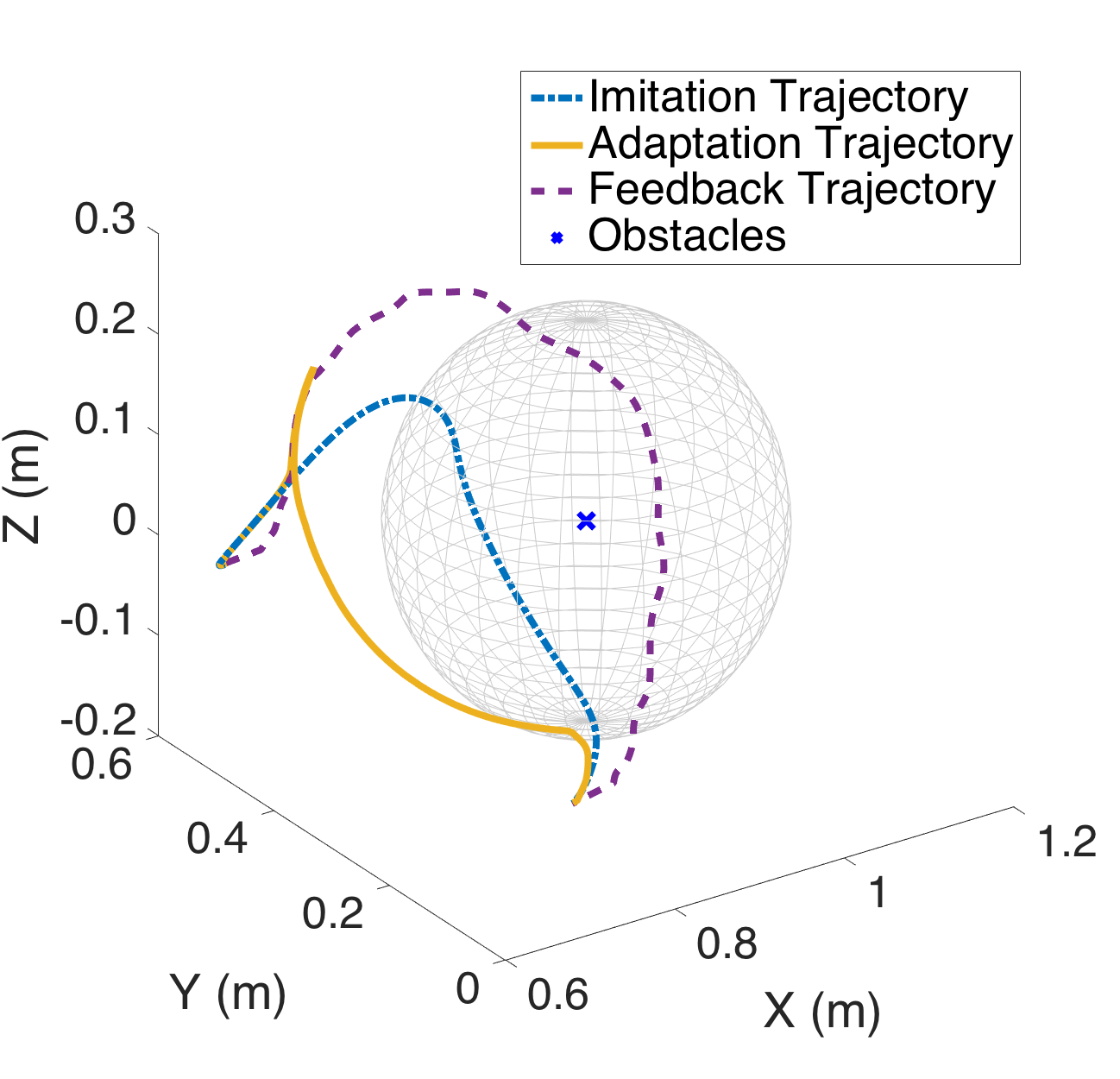} 
    \label{fig:duckadaptation}
  }
  \subfigure[]{
    \centering
    \includegraphics[height=1.45in]{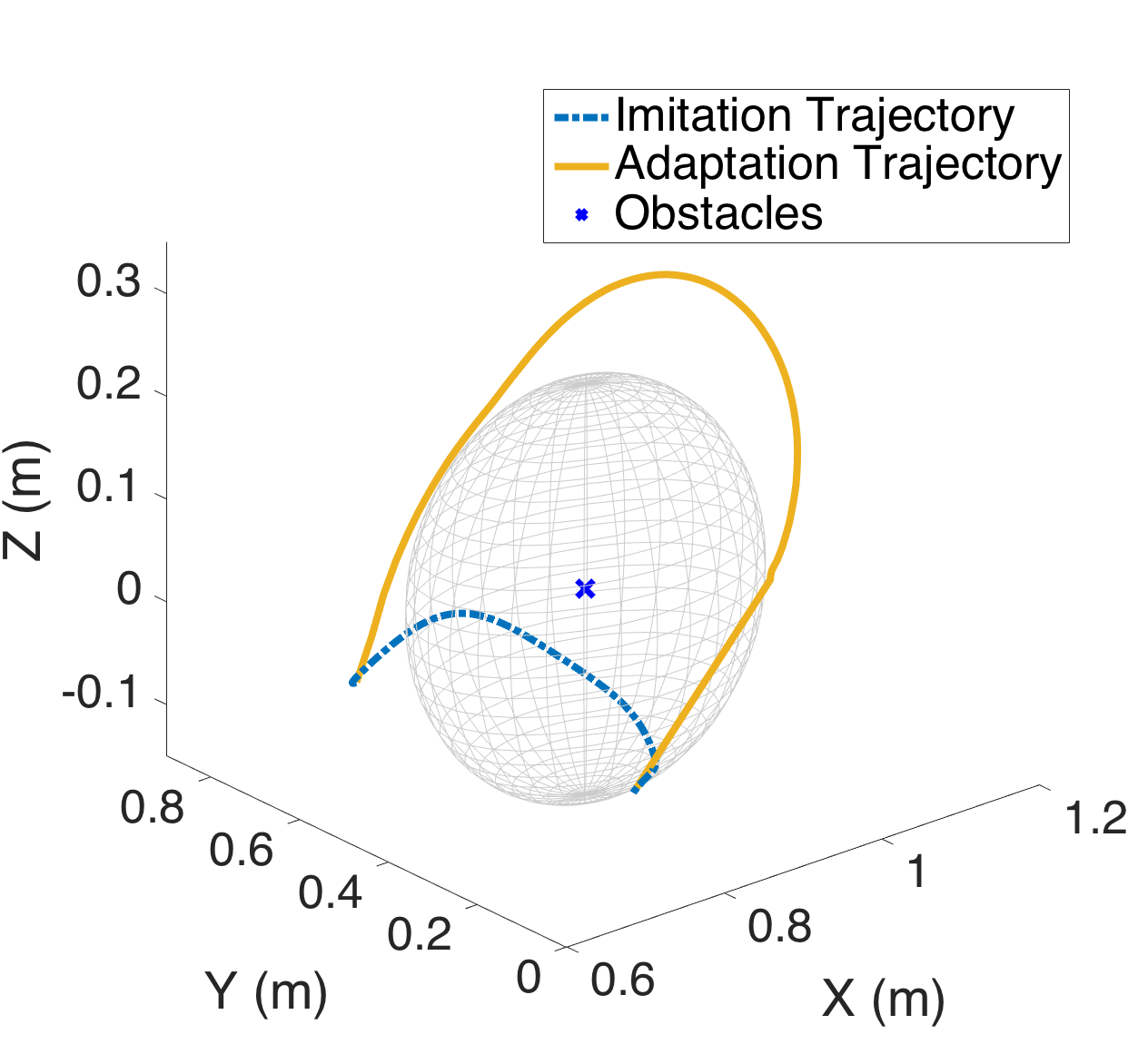} 
    \label{fig:duckimproved}
  }
  \label{fig:duckexperiment}
  \caption{Baxter Learning to Adapt Movement for Transferring Knife: (a) (c) Movement adaptation with initial weights, successfully avoided duck doll around it but may risk scratches, afterwards feedback trajectory is provided for adaptation preferences; (b) (d) Movement adaptation for different situation, with updated weights after learning from feedback trajectory, successfully avoided the duck doll above it as desired.}
  \vspace{-10pt}
\end{figure}

In a second scenario where a robot is transferring a knife around some fragile obstacle, the user may prefer robot to avoid the obstacle above it instead of around it. With the same methods here, we could also generate adapted trajectories as shown in \figref{fig:baxterduckadaptation} and \figref{fig:duckadaptation} for initial weights. With the user provided feedback trajectory, the robot successfully learns the user specified preferences for movement adaptation and generates the improved adapted trajectories for different situations as shown in \figref{fig:baxterduckimproved} and \figref{fig:duckimproved}.

\section{Conclusion and Future Work}
We present a framework for learning to adapt robot end effector movement for manipulation tasks. The proposed method generalizes offline learned movement skills to novel situations considering obstacle avoidance and other task-dependent constraints. It adapts the imitation trajectory generated from demonstrations, while maintaining the learned movement pattern and considering the variations, to avoid obstacles with desired directions and distances and keep a safety margin within a workspace. Also it provides a way to learn how to adapt the movement by on-line interactions from user's feedback. 

Besides learning how to adapt movement from user's feedback, the visual information of the objects and the environment could also indicate the preferences of movement adaptation. For instance, the deviation direction for avoiding a knife could be inferred directly from the location of its blade from visual space. We are further investigating the possibility of directly learning the preferences to adapt movement from visual perception of the task context.

\addtolength{\textheight}{-12cm}   









\bibliographystyle{IEEEtran}
\bibliography{root}

\end{document}